\documentclass{article}
\usepackage{AGA_GI}

\usepackage{times}

\usepackage{soul}
\usepackage{url}
\usepackage{graphicx}
\usepackage{subcaption}
\usepackage{multirow}
\usepackage{amssymb}
\usepackage[flushleft]{threeparttable}
\usepackage{amsmath}
\usepackage{booktabs}
\newcommand{\rulesep}{\unskip\ \vrule\ }

\title{Incorporating Effective Global Information via Adaptive Gate Attention \\for Text Classification}

\author{
Xianming Li$^1$\footnote{Contact Author}\and
Zongxi Li$^2$\and
Yingbin Zhao$^1$\and
Haoran Xie$^3$ \And
Qing Li$^4$\\
\affiliations
$^1$AI Department, Ant Financial Services Group\\
$^2$Department of Computer Science, City University of Hong Kong\\
$^3$Department of Computing and Decision Sciences, Lingnan University\\
$^4$Department of Computing, Hong Kong Polytechnic University \\
\emails
\{niming.lxm, zyb166123\}@antfin.com,
zongxili2-c@my.cityu.edu.hk,
hrxie@ln.edu.hk,
csqli@comp.polyu.edu.hk
}

\begin{document}

\maketitle

\begin{abstract}
The dominant text classification studies focus on training classifiers using textual instances only or introducing external knowledge (e.g., hand-craft features and domain expert knowledge). In contrast, some corpus-level statistical features, like word frequency and distribution, are not well exploited. Our work shows that such simple statistical information can enhance classification performance both efficiently and significantly compared with several baseline models. In this paper, we propose a classifier with gate mechanism named Adaptive Gate Attention model with Global Information (AGA+GI), in which the adaptive gate mechanism incorporates global statistical features into latent semantic features and the attention layer captures dependency relationship within the sentence. To alleviate the overfitting issue, we propose a novel Leaky Dropout mechanism to improve generalization ability and performance stability. Our experiments show that the proposed method can achieve better accuracy than CNN-based and RNN-based approaches without global information on several benchmarks.

\end{abstract}

\section{Introduction}
Text classification is playing an essential role in Natural Language Processing (NLP) as one of the fundamental tasks with broad applications. The mainstream deep text classifiers suffer from the data sparseness issue, and to enrich semantic features, researchers turn to some useful external knowledge as complementary information, i.e., tags, character, POS, sentiment lexicon, entity knowledge base. Their studies show that introducing proper external knowledge is helpful to the classification task. However, we notice that some most primitive features are overlooked in the deep learning era, i.e. word frequency and distribution, which are fixed and intrinsic feature of a corpus. The most representative algorithm utilizing statistical feature is the \emph{term frequency-inverse document frequency} (TFIDF), which is a straightforward information retrieval technique for document modelling. However, because of the bag-of-word nature, TFIDF is unable to utilize positional information and capture fine-grained semantics, which makes it a less favourable feature compared with word embeddings in the deep architecture. Nevertheless, to our surprise, we find that using term-count-of-labels statistics (defined in Section \ref{definition}) as an auxiliary feature shows substantial improvements in the pilot study, in which the word frequency adapts weights of terms via a simple attention layer. We believe that the researchers may underestimate the real power of global statistics feature in deep learning; therefore, in this work, we design a new framework to fuse such features elegantly. 

When designing the fusion mechanism, we think of two major concerns: 1. The semantic feature and statistical feature are not compatible in scale and dimension; 2. The new information may be not necessary for all semantic features. 
To address the first concern, we employ non-linear projection to map both features into a shared information space to make both latent representations compatible with each other. The second concern raises a new perspective towards the method of using \emph{additional information}\footnote{We use the term \emph{addition information} to denote both statistical feature and external knowledge, which are additional to the semantic features, in the remaining part of this paper for concise expression.}. We argue that not every semantic feature need to be enhanced since some additional information may introduce noise to the classifier. Therefore, instead of element-wise operation, we design an Adaptive Gate module to add auxiliary information to the less-confident semantic features only (with values around $0.5$ after sigmoid activation) while the high-confident semantic features remain unchanged. By doing this, the proposed model can achieve a balance between the original semantic features and the additional features for better decision making.

Moreover, we notice that the current Dropout mechanism may not be compatible with the proposed feature fusion mechanism. Neurons enriched by additional information and neurons containing preserved semantic features may be deactivated during training, and the improvement brought by the new architecture can be offset. Therefore, instead of directly deactivating neurons, we propose a novel Leaky Dropout mechanism to reduce the value of selected neurons only and conduct further experiments to demonstrate the effectiveness of this novel mechanism.


The main contributions of this paper are summarized as
follows:
\begin{itemize}
    \item We leverage corpus-level statistics feature to enrich semantic features for text classification. To retrieve necessary and useful global information only, we propose a well-designed adaptive gate module with attention mechanism to fuse statistics feature into semantic features with low confidence. 
    \item We propose a novel Leaky Dropout to alleviate overfitting issue without fully deactivating neurons, which is demonstrated to be more robust than conventional Dropout.
    \item We conduct extensive experiments on six small-scale datasets and two large-scale datasets with significance test. The results show that our models significantly outperform state-of-the-art methods.
\end{itemize}

\section{Methodology}
\subsection{Global information}
\label{definition}
We first formally define the adopted global information as follows.\\
\textbf{Definition 1} \emph{Term-count-of-labels} (TCoL) is a global statistics of a term towards the labels. Given a word $w$ and a set of labels of $c$ classes, the TCoL vector $w$ is 
\begin{equation}
    \boldsymbol{\zeta}^w = [\zeta_1, \dots, \zeta_c ],
\end{equation} 
where $\zeta_i$ is the count of word $w$ on label $i$. Given a sentence $s = \{w_i\}^c_{i=1}$, the TCoL matrix of sentence $s$ is 
\begin{equation}
    \boldsymbol{\zeta}^s=[\boldsymbol{\zeta}^{w_1},\dots, \boldsymbol{\zeta}^{w_c}].
\end{equation}\par
The term-count-of-labels describes the global distribution of labels as a feature of the word. Such features are primitive but highly informative in feature selection. Intuitively, if a word has very high or very low frequency on all labels, then we shall assume this word has limited contribution to the classification task. In contrast, if a word appears more frequently in one class, we assume this word is more informative and hope our classifier can highlight this word in decision making. The TCoL dictionary is obtained only from the training set. 


\subsection{Adaptive Gate Attention Network}
\begin{figure*}
    \centering
    \includegraphics[width = .95\textwidth]{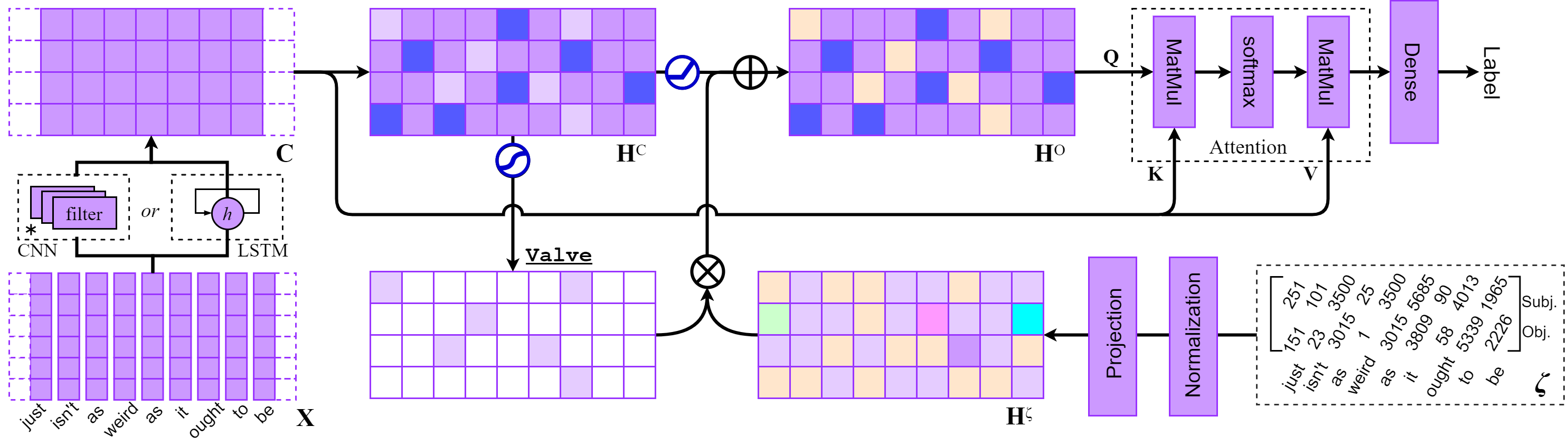}
    \caption{Generic framework. AGA-CNN+GI and AGA-LSTM+GI are two variants of AGA model with CNN and LSTM as feature extractor respectively. The \emph{subj} and \emph{obj} are labels of \textbf{Subj} dataset.}
    \label{fig_generic}
\end{figure*}

Figure \ref{fig_generic} shows a generic framework of the proposed model, consisting of input layer, feature-extraction layer, adaptive gate attention (AGA) module and output layer. Since there exists different feature-extraction techniques, we implement AGA-CNN and AGA-LSTM as two variants of the model employing CNN and LSTM as feature extractor respectively. Given an annotated documents, the AGA+GI model firstly maps the document into embedding vector matrix and deploy convolutional or recursive operation to obtain latent semantic representations. Then, a well-designed AG module projects semantic representation and TCoL matrix to shared information space. It fuses global statistic features into the pipeline by selectively combining units from both matrices, in which the global information makes semantic representation more informative. Finally, the model extracts higher-level features utilizing attention layers and forwards the features in fully-connected layers to output logits for label prediction and loss calculation. During the training, we compute cross-entropy loss and employ ADAM optimizer to train the classifier.




\subsubsection{Input layer}
The input of model is a sentence with fixed length $m$ and the TCoL matrix $\boldsymbol{\zeta}$ of the sentence. we first map each word into a $k$-dimensional continuous space and obtain the word embedding vector $\mathbf{x}_m \in \mathbb{R}^k$. Then we concatenate all word vectors to form a $k \times m$ matrix as model input:
\begin{equation}
    \mathbf{x} = [\mathbf{x}_1, \mathbf{x}_2, \dots, \mathbf{x}_m]
\end{equation}
We pad the sentences to keep a uniform length for all sentences following the same way in \cite{kim2014convolutional}. 

\subsubsection{Feature extraction layer}
We employ convolutional operation or recurrent operation to produce latent feature map. \par
For a CNN layer, we apply a filter $\mathbf{W}^f \in \mathbb{R}^{h\times k}$ with window size $h$. The new feature $c_i$ is generated from a window of word vectors $\mathbf{x}_{i:i-h+1}$:
\begin{equation}
    c_i = \boldsymbol{f}(\mathbf{W}_f \circledast \mathbf{x}_{i:i-h+1} + b),
\end{equation}
here, $b \in \mathbb{R}$ is the bias term, and $\boldsymbol{f}(\cdot)$ is a non-linear function. Each filter produces a feature vector $\mathbf{c} = [c_1, c_2, \dots, c_{m}]^\intercal$ with padding. We employ $d$ filters in this layer to produce a latent feature map $\mathbf{C} \in \mathbb{R}^{d \times m}$ in the semantic space. 

For an LSTM layer, we follow the same settings with \cite{hochreiter1997} and \cite{zhou2015c}. At time $t$, the latent feature $\mathbf{c}_t$ in each LSTM cell is obtained as follow,
\begin{equation}
    \mathbf{c}_t = \mathrm{LSTM}(\mathbf{c}_{t-1}, \mathbf{x}_t).
\end{equation}
With hidden dimension set to $d$, we have the whole semantic feature $\mathbf{C} \in \mathbb{R}^{d \times m}$.


\subsubsection{Adaptive Gate Attention module}
To merge global information with semantic features, we map the semantic feature map $\mathbf{C}$ and the normalized TCoL matrix $\boldsymbol{\zeta}^s$ to a shared information space $\mathbf{H} \in \mathbb{R}^{d \times m}$ through the fully-connected layers as follow,
\begin{equation}
\begin{split}
    \mathbf{H}^{C} &= \mathbf{W}^{C} \cdot \mathbf{C} + \mathbf{b}^{C} \\
    \mathbf{H}^{\zeta} &= \mathbf{W}^{\zeta} \cdot (\boldsymbol{\zeta}^s / V) + \mathbf{b}^{\zeta}
\end{split}
\end{equation}
where $\mathbf{H}^{C}$ and $\mathbf{H}^{\zeta}$ are latent representations of semantic feature and term frequency respectively in the information space, and $\boldsymbol{\zeta}^s$ is normalized by dividing the length of the dictionary $V$. We find this projection is very crucial in Section \ref{sec_results}.

The AG module fuses $\mathbf{H}^{C}$ and $\mathbf{H}^{\zeta}$ to output a GI-enhanced feature map $\mathbf{H}^O$ through the $\mathbf{AdaGate}$ function,
\begin{equation}
\begin{split}
     \mathbf{H}^O &= \mathbf{AdaGate} (\mathbf{H}^{C}, \mathbf{H}^{\zeta}, \epsilon) \\
     &= \mathrm{ReLU}(\mathbf{H}^{C}) +  \mathbf{Valve} (\sigma(\mathbf{H}^{C}), \epsilon) \odot \mathbf{H}^{\zeta},
\end{split}
\end{equation}
where $\mathrm{ReLU}(\cdot)$ and $\sigma(\cdot)$ are activation functions, and $\odot$ stands for element-wise production. The $\sigma(\cdot)$ function produces values in probability form, and the $\mathbf{Valve}$ function is designed to restore less-confident entries (with probability near $0.5$) for matching with elements in $\mathbf{H}^{\zeta}$. More concretely, for every unit $a \in \mathbf{H}^{C}$, 
\begin{equation}
    \mathbf{Valve}(a, \epsilon) =
\begin{cases}
    a,  & \text{if  } 0.5-\epsilon \leq a \leq 0.5 + \epsilon\\
    0, & \text{otherwise}
\end{cases}
\end{equation}
where $\epsilon$ is a leaky hyper-parameter tuning the threshold of confidence, specifically, we dump all statistical information if $\epsilon = 0$, and combine statistical information with all semantic features if $\epsilon = 0.5$. Therefore, the element-wise production exploits $\mathbf{Valve} (\sigma(\mathbf{H}^{C}), \epsilon)$ as a filter to extract necessary global information only, thus $\mathbf{H}^O$ can be more informative by restoring both essential semantic features and additional GI features.\par
After fusing GI, we normalize $\mathbf{H}^O$ via $\mathbf{softmax}$ to get attention weight $\boldsymbol{\alpha}$,
\begin{equation}
    \boldsymbol{\alpha}_i = \frac{\exp{(\mathbf{H}^O_i)}}{\sum^{m}_{j=1} \exp{(\mathbf{H}^O_j)}}.
\end{equation}
Then we apply attention weight to the semantic representation $\mathbf{C}$ to produce feature vector $\mathbf{a}$, 
\begin{equation}
    \mathbf{a} = \sum^{m}_i \boldsymbol{\alpha}_i \cdot \mathbf{C}_i.
\end{equation}

\subsubsection{Output layer \& loss function}
After passing through fully-connected layers with Leaky Dropout (discussed in Section \ref{leaky_dropout}) and softmax layer, feature vector $\mathbf{a}$ is mapped to the label space for label prediction and loss calculation. To maximize the probability of the correct label $y$, we deploy optimizer to minimize cross-entropy loss $L$, which is defined as
\begin{equation} 
L(\mathbf{a}, y) = - \frac{1}{N} \left[ \sum_i^N \sum_j^c \mathbf{1} (y_i = j) \ln \frac{\exp{(\mathbf{a}_j^{(i)})}}{\sum_t \exp{(\mathbf{a}_t^{(i)})}} \right].
\end{equation}

\subsection{Leaky Dropout}
\label{leaky_dropout}
\begin{figure}[ht]
    \centering
    \begin{subfigure}{.37\columnwidth}
    \includegraphics[width=1\columnwidth]{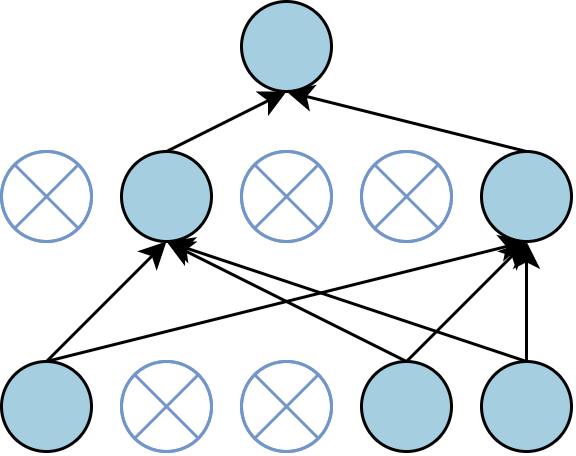}
    \subcaption{Dropout}
    \label{fig_dropout}
    \end{subfigure}
    \rulesep
    \begin{subfigure}{.37\columnwidth}
    \includegraphics[width=1\columnwidth]{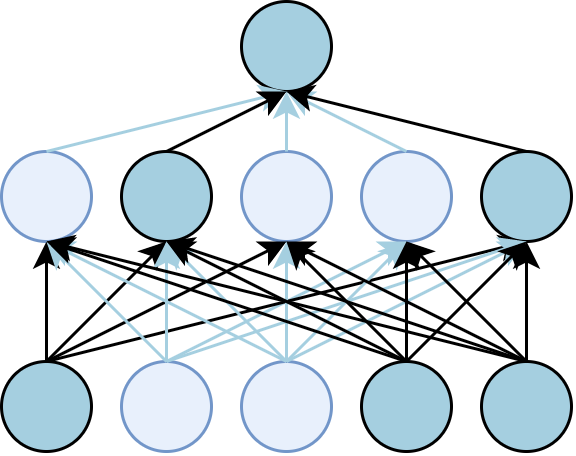}
    \subcaption{Leaky Dropout}
    \label{fig_ldropout}
    \end{subfigure}
    \caption{Comparison between conventional Dropout mechanism and the proposed Leaky Dropout.}
\end{figure}

The Dropout \cite{srivastava2014dropout} is an elegant mechanism to alleviate overfitting issue when training a deeper network as shown in Figure \ref{fig_dropout}, but directly blocking a selected neuron may have some unexpected side effects. Therefore, we propose a soft mechanism called Leaky Dropout to suppress the weights of selected neurons rather than completely deactivate them, as shown in Figure \ref{fig_ldropout}. Given an input feature vector $\mathbf{x}$ and dropout rate $\beta$, the partially suppressed vector $\mathbf{x'}$ is obtained by computing element-wise production of $\mathbf{x}$ and a mask vector $\mathbf{m}$:
\begin{equation}
    \mathbf{x'} = \mathbf{x} \odot \mathbf{m},
\end{equation}
and the value of each element in $\mathbf{m}$ is assigned as:
\begin{equation}
\begin{split}
    \mathbf{m}_i &=
\begin{cases}
    \frac{1}{1-\beta},  & z_i = 1\\
    \gamma, & z_i = 0
\end{cases}
\\ z_i &\sim Bernoulli(z)
\\ \gamma  &= \frac{1-\beta}{c^2},
\end{split}
\end{equation}
where $z_i$ is sampled from a Bernoulli distribution as indicators (preserve if $z = 1$ and suppress if $z = 0$), and parameter $c$ is set to control how hard the suppression will be. Following the same setting in \cite{srivastava2014dropout}, we magnify preserved cells to keep expectations unchanged.

\section{Experiment}
\begin{table}[ht]
    \centering
    \begin{threeparttable}
    \begin{tabular}{l||c|c|c|c|c}
    \toprule
         Data& c & l & N & V & Test  \\
         \midrule
         CR & $2$ & $20$ & $10,662$ & $18,765$ & CV \\
         Subj & $2$ & $23$ & $10,000$ & $21,323$ & CV \\
         SST-1 & $5$ & $18$ & $11,855$ & $17,836$ &  $2,210$ \\
         SST-2 & $2$ & $19$ & $9,613$ & $16,185$ & $1,821$ \\
         TREC & $6$ & $10$ & $5,952$ & $9,592$ & $9,125$ \\
         MPQA & $2$ & $3$ & $10,606$ & $6,246$ & CV \\
         Yelp F. & $5$ & $596$ & $1m^{*}$ & $313,669$ & $100k$ \\
         Yelp P. & $2$ & $585$ & $1m^{*}$ & $311,400$ & $100k$ \\
         \bottomrule
    \end{tabular}
    \begin{tablenotes}
    \small
    \item \text{*} We sample 1 million data from both datasets.
    \end{tablenotes}
    
    \end{threeparttable}
    \caption{Summary statistics for the datasets. $c$: Number of classes. $l$: Average sentence length. $N$: Dataset size. $V$: Vocabulary size. $Test$: Size of testset (CV means no standard train/test split thus we deploy cross-validation).}
    \label{tab-dataset}
    
\end{table}

\begin{table*}[ht]
    \centering
    \begin{threeparttable}
    \begin{tabular}{llllllll}
    \toprule
        \multirow{2}{*}{\textbf{Model}}  & \multicolumn{3}{c}{F1 (\%)} & &  \multicolumn{3}{c}{Accu. (\%)} \\
        \cmidrule{2-4} \cmidrule{6-8}
        &  \multicolumn{1}{c}{SST-1} & \multicolumn{1}{c}{TREC} & \multicolumn{1}{c}{Yelp F.} & & \multicolumn{1}{c}{SST-1} & \multicolumn{1}{c}{TREC} & \multicolumn{1}{c}{Yelp F.}  \\
        \midrule
        SVM & $37.48^{\mathsection}\pm .00$ &  $87.80^{\mathsection}\pm .00$ &  \multicolumn{1}{c}{$-$} & & $39.36^{\mathsection}\pm .00$ & $87.52^{\mathsection} \pm .00$ & \multicolumn{1}{c}{$-$} \\
        
        TextCNN & $34.69^{\mathsection} \pm .56$   & $90.55^{\mathsection}\pm .74$ & $63.82^{\mathsection} \pm .26$ & & $44.96^{\mathsection}\pm .81$ & $90.68^{\dagger} \pm .57$ & $67.25^{\mathsection} \pm .19$ \\
        
        DPCNN & $39.16^{\mathsection} \pm .57$ & $91.82^{\;\,} \pm .91$ & $66.97 ^{\mathsection}\pm .22$ & &$45.80^{\dagger} \pm .76$ & $91.76^{\;\,} \pm .96$ & $68.90^{\mathsection} \pm .34$ \\
        
        Bi-LSTM & $43.14^{\;\,} \pm .58$ &$92.13^{\;\,}\pm .73$ & $69.50^{\dagger}\pm .14$ & & $46.03^{\dagger}\pm .68$ & $92.19^{\;\,}\pm .71$ &  $70.31^{\mathsection}\pm .71$   \\
        C-LSTM & $41.69^{\mathsection} \pm .65$ & $92.33^{\;\,} \pm .41$ & $69.61^{\dagger}\pm .33$ & & $45.82^{\dagger} \pm .83$ & $92.23^{\;\,}\pm .51$ & $70.48^{\dagger}\pm .22$  \\
        
        FastText & $35.94^{\mathsection}\pm .58$ &  $85.91^{\mathsection}\pm .28$ &  $59.27^{\mathsection}\pm .05$ & & $41.12^{\mathsection}\pm .96$ & $86.18^{\mathsection} \pm .30$ & $68.38^{\mathsection}\pm .13$  \\
        
        \midrule
        AGA-CNN w/o GI & $39.92^{\;\,}\pm .64$ &  $92.30^{\;\,}\pm .70$ &  $65.40^{\;\,}\pm .42$ & & $46.57^{\;\,}\pm .57$ & $92.39^{\;\,}\pm .78$ & $69.00^{\;\,}\pm .22$ \\
        AGA-LSTM w/o GI & $\mathbf{43.59}^{\!\;}\pm .51$ & $91.03^{\;\,}\pm .50$ &  $69.44^{\;\,}\pm .21$ & &  $46.20^{\;\,} \pm .49$ & $90.94^{\;\,}\pm .58$ & $70.48^{\;\,} \pm .11$  \\
        AGA-CNN w/ GI & $39.93^{\;\,}\pm .45^{\ast}$ &  $\mathbf{92.42}^{\!\;}\pm .77^{\ast}$ &  $65.99^{\;\,}\pm .15^{\ast\ast}$ & & $\mathbf{46.64}^{\!\;}\pm .64^{\ast}$ & $\mathbf{92.55}^{\!\;}\pm .73^{\ast}$ & $69.19^{\;\,}\pm .30^{\ast\ast}$ \\
        AGA-LSTM w/ GI & $43.21^{\;\,}\pm .66^{\ast}$ & $91.40^{\;\,}\pm .54^{\ast}$ &  $\mathbf{69.71}^{\!\;}\pm .11^{\ast\ast}$ & &  $46.03^{\;\,} \pm .52^{\ast}$ & $91.45^{\;\,}\pm .36^{\ast}$ & $\mathbf{70.71} ^{\!\;}\pm .14^{\ast\ast}$  \\

    \bottomrule
    \end{tabular}
    \begin{tablenotes}
    \small
    \item $^{\dagger} p<.05$, $^{\dagger} p<.01$, $^{\mathsection} p<.001$.  $^{\ast} \epsilon=0.05$, $^{\ast\ast} \epsilon=0.25$.
    \end{tablenotes}
    \end{threeparttable}
    \caption{Results on Multi-class datasets. We evaluate the performance using F1 score and Accuracy. Baseline results with $p$-value indicated means our methods have significant improvement compared with this baseline. $\epsilon$ is the leaky parameter in the adaptive gate.}
    \label{tab_results_milti}
\end{table*}

\begin{table*}[ht]
    \centering
    \begin{threeparttable}
    \begin{tabular}{llllll}
    \toprule
        \multirow{2}{*}{\textbf{Model}}  & \multicolumn{5}{c}{Accu. (\%)} \\
        \cmidrule{2-6}
         & \multicolumn{1}{c}{CR} & \multicolumn{1}{c}{Subj} & \multicolumn{1}{c}{SST-2} & \multicolumn{1}{c}{MPQA} & \multicolumn{1}{c}{Yelp P.}  \\
        \midrule
        SVM & $79.34^{\mathsection}\pm 1.72$ & $90.90^{\mathsection}\pm .71$ & $81.44^{\mathsection}\pm .00$ & $86.29^{\mathsection}\pm 1.15$ & \multicolumn{1}{c}{$-$}  \\
        TextCNN & $82.25^{\mathsection}\pm 1.40$  & $93.16^{\mathsection}\pm .63$ & $84.20^{\mathsection}\pm .29$ & $89.67^{\dagger}\pm 1.05$ & $94.23^{\mathsection} \pm .20$   \\
        DPCNN & $83.66^{\;\,}\pm 1.69$ &  $93.61^{\dagger}\pm .70$ & $84.46^{\dagger}\pm .36$ & $89.83^{\;\,}\pm 1.23$ & $95.31^{\mathsection}\pm .13$ \\
        Bi-LSTM & $83.18^{\dagger}\pm 1.64$ & $94.02^{\;\,}\pm .76$ & $84.64^{\dagger}\pm .23$ & $89.64^{\;\,}\pm 1.18$ & $95.29^{\mathsection}\pm .20$  \\
        C-LSTM & $83.28^{\;\,}\pm 1.47$ & $93.99^{\;\,}\pm .48$ & $84.64^{\dagger}\pm .28$ & $89.49^{\dagger}\pm 1.31$ & $95.53^{\mathsection}\pm .08$   \\
        FastText & $79.08^{\mathsection}\pm 2.04$ & $90.72^{\mathsection}\pm .67$ & $81.31^{\mathsection}\pm .38$ & $85.76^{\mathsection}\pm 1.41$ & $95.52^{\mathsection}\pm .04$  \\
        \midrule
        AGA-CNN w/o GI & $83.75^{\;\,}\pm 1.61$ & $93.40^{\;\,} \pm .80$ & $84.47^{\;\,} \pm .37$ & $89.81^{\;\,}\pm 1.03$ & $96.01^{\;\,}\pm .06$   \\
        AGA-LSTM w/o GI & $83.65^{\;\,}\pm 1.35$ & $93.99^{\;\,}\pm .72$ & $84.74^{\;\,}\pm .31$ & $89.61^{\;\,}\pm 1.09$ & $95.36^{\;\,}\pm .05$   \\
        
        AGA-CNN w/ GI & $\mathbf{84.18 }^{\!\;}\pm 1.36 ^{\ast}$ & $93.69^{\;\,} \pm .54^{\ast}$ & $84.73^{\;\,} \pm .41^{\ast}$ & $\mathbf{89.91}^{\!\;}\pm 1.21^{\ast}$ & $96.12^{\;\,}\pm .30^{\ast\ast}$   \\
        AGA-LSTM w/ GI & $83.72^{\;\,}\pm 1.63^{\ast}$ & $\mathbf{94.22}^{\!\;}\pm .70^{\ast}$ & $\mathbf{85.11}^{\!\;}\pm .37^{\ast}$ & $89.37^{\;\,}\pm 1.21^{\ast}$ & $\mathbf{96.57}^{\!\;}\pm .04^{\ast\ast}$   \\
    \bottomrule
    \end{tabular}
    \begin{tablenotes}
    \small
    \item $^{\dagger} p<.05$, $^{\dagger} p<.01$, $^{\mathsection} p<.001$.  $^{\ast} \epsilon=0.05$, $^{\ast\ast} \epsilon=0.25$.
    \end{tablenotes}
    \end{threeparttable}
    \caption{Results on Binary datasets. We evaluate the performance using Accuracy (also equal to F1 score). Baseline results with $p$-value indicated means our methods have significant improvement compared with this baseline.  $\epsilon$ is the leaky parameter in the adaptive gate.}
    \label{tab_results_bin}
\end{table*}

\subsection{Datasets}
We test the proposed model on various datasets. \textbf{CR} \cite{hu2004mining} contains customer reviewers of various products with reviews annotated with positive or negative. \textbf{Subj} \cite{pang2004sentimental} is a dataset labeled with sentence subjectivity. Each sentence is annotated with subjective or objective. \textbf{SST-1} \cite{socher2013recursive} (Stanford Sentiment TreeBank) is a dataset of movie reviews with five fine-grained sentiment labels (i.e. very positive/negative, positive/negative, neutral). This dataset has a standard train/dev/test split. \textbf{SST-2} \cite{socher2013recursive} is Stanford Sentiment TreeBank dataset with binary sentiment labels. \textbf{TREC}  \cite{li2002learning} is a question dataset with questions of six types about person, location, numeric information, etc. \textbf{MPQA}  \cite{wiebe2005annotating} is the opinion polarity detection subtask of the MPQA dataset. \textbf{Yelp Review Full} (Yelp F.) is the reviews subset of Yelp Open Dataset consists of sentences with polarity star labels ranging from 1 to 5. \textbf{Yelp Review Polarity} (Yelp P.) is the reviews subset of Yelp Open Dataset. Compared with Yelp F., Yelp P. only has binary labels (negative and positive). Summary statistics of the datasets are shown in Table \ref{tab-dataset}. \par
We deploy 10-fold cross-validation in the datasets without standard train/test split. Due to the limitation of computation power, we randomly sample 1 million data from Yelp F. and Yelp P. and divide sampled data into train/test sets.

\subsection{Baselines}
We compare the proposed model with following text classifiers. To evaluate the contribution of frequency information explicitly, we use TFIDF as features and apply \textbf{SVM} as classifier on small datasets. \textbf{TextCNN} \cite{kim2014convolutional} is a popular CNN-based classifier exploiting one-dimensional convolution operation on embedding matrix and max-over-time pooling on extracted the feature map. \textbf{DPCNN} \cite{johnson-zhang-2017-deep} is a low-complexity word-level deep CNN model in pyramid shape employing downsampling module and shortcut connections. \textbf{Bi-LSTM} \cite{graves2013hybrid} is a bi-directional LSTM model extracting both forward and reverse sequential features. \textbf{C-LSTM} \cite{zhou2015c} employs CNN model to extract a sequence of higher-level semantic features and feeds these vectors into an LSTM network to obtain the sentence representation for classification. \textbf{FastText} \cite{joulin-etal-2017-bag} treats the average of word/n-grams embeddings as document embeddings, then feeds document embeddings into a linear classifier. \textbf{AGA-CNN w/o GI} and \textbf{AGA-LSTM w/o GI} are baselines for ablation study with $\epsilon$ set to $0$, which means the gate rejects all additional information, but we preserve the projection $\mathbf{C} \mapsto \mathbf{H}^C$.

\subsection{Implementation details}

\subsubsection{Evaluation metrics}
We evaluate the model performance and the significance of improvement using the following metrics. \textbf{F1 score} measures both precision and recall as a whole. We report \emph{Macro}-average results on both multi-class and binary-class datasets in this paper. \textbf{Accuracy} measures how many instances are correctly classified among all instances. \textbf{T-test} reveals how significant the improvements are and we report $p$-value of the proposed model compared with baselines for each trail. 
    
\subsubsection{Word embedding}
It is a widely adopted approach to improve model performance by initializing word vectors with pre-trained language model. We adopt the publicly available \emph{FastText} \footnote{https://fasttext.cc/docs/en/english-vectors.html} \cite{mikolov2018advances}, which has 1 million word vectors trained on Wikipedia 2017, UMBC webbase corpus and statmt.org news dataset (16B tokens) with the dimensionality of 300. Words not present in the pre-trained model are initialized randomly.

\subsubsection{Parameter settings}
The parameters involved in all CNN and RNN modules follow the settings in their original papers. More concretely, the CNN-based models have filter size of $[3, 4, 5]$ with $100$ filters of each, and the RNN-based models have hidden dimension of $128$. All models adopt Adam optimizer with batch size of $64$ and dropout rate of $0.5$ .

\subsection{Results \& Discussion}
\label{sec_results}
Results of our models against other methods and results of ablation study are listed in Table \ref{tab_results_milti} (multi-class datasets) and Table \ref{tab_results_bin} (Binary-class datasets) respectively. In general, the proposed model achieves the best accuracy on all datasets and best F1 score on most datasets (except TREC). The t-test indicates the proposed methods have significant improvement in the majority of all results ($50$ out of $63$). We conduct additional experiments to validate the effectiveness of the proposed Leaky Dropout and show the results in Figure \ref{fig_accu_dropout}.

\subsubsection{Effect of AGA module}
We had initially designed the projection layer and the attention layer of AGA module as a designated mechanism to incorporate GI. However, as shown in Table \ref{tab_results_milti} and \ref{tab_results_bin}, the AGA module can improve the performance of CNN-based model significantly on all datasets even without GI. Specifically, we see improvements of $5.23\%$ on F1 of SST-1, $2.17\%$ on F1 of Yelp F. and $1.78\%$ on accuracy of Yelp P., suggesting the proposed method is more potent on semantic feature selection than the original max-over-time pooling method. Besides, compared with all baseline models, models with AGA module have achieved significant improvements on dataset Yelp F. and Yelp P., which indicates that AGA module may be more suitable to deal with complicated dependency relationship when the data amount is enormous.

\subsubsection{Effect of GI}
The contribution of GI is distinctly identified as models with GI achieve the best performance on most datasets in the ablation experiments. We also remark that the improvements brought forth by GI can be affected by the leaky constant $\epsilon$ in the gate module, which controls the confidence interval to trigger the information fusion. On large datasets, a larger $\epsilon$ can produce higher accuracy than a smaller $\epsilon$, while on small datasets, a smaller $\epsilon$ is more favourable ($0.05$ in this case), which possibly due to the bias from TCoL information on small datasets. As defined previously, TCoL is a frequency statistic of words towards labels, which can be easily deviated from the real distribution by the limited size of a dataset. To deal with such a problem, a tighter interval is preferred in order to make the model depend more semantic features, which can alleviate the influence of the bias. In contrast, for large datasets, the TCoL statistics can approximate the real distribution and be more helpful to the model training.
\begin{figure}[ht]
    \centering
    \begin{subfigure}{.45\columnwidth} 
    \includegraphics[width=1\columnwidth]{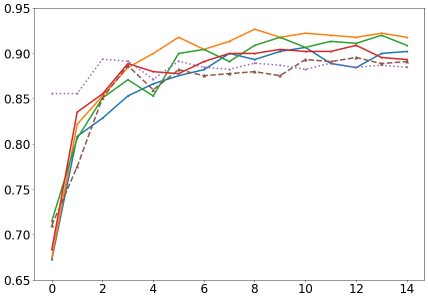}
    \subcaption{Accu. of AGA-CNN}
    \label{fig_a}
    \end{subfigure}
    \begin{subfigure}{.45\columnwidth}
    \includegraphics[width=1\columnwidth]{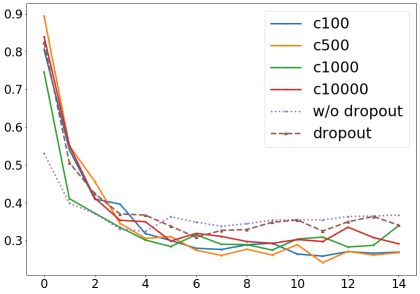}
    \subcaption{Loss of AGA-CNN}
    \end{subfigure}

    \begin{subfigure}{.45\columnwidth} 
    \includegraphics[width=1\columnwidth]{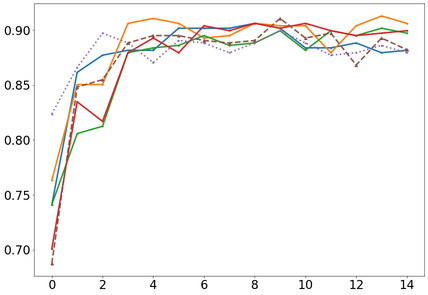}
    \subcaption{Accu. of AGA-LSTM}
    \label{fig_c}
    \end{subfigure}
    \begin{subfigure}{.45\columnwidth}
    \includegraphics[width=1\columnwidth]{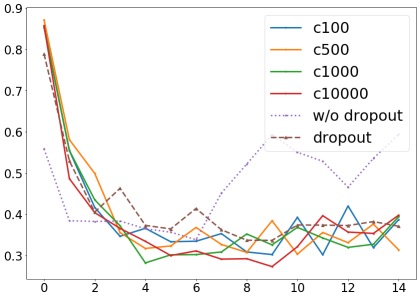}
    \subcaption{Loss of AGA-LSTM}
    \end{subfigure}    
    
    \begin{subfigure}{.45\columnwidth}
    \includegraphics[width=1\columnwidth]{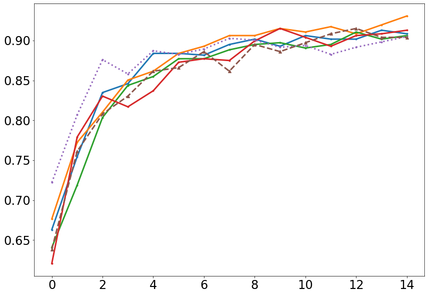}
    \subcaption{Accu. of TextCNN}
    \end{subfigure}
    \begin{subfigure}{.45\columnwidth}
    \includegraphics[width=1\columnwidth]{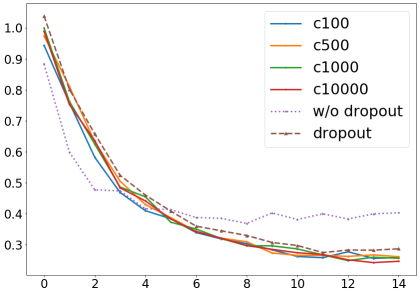}
    \subcaption{Loss of TextCNN}
    \end{subfigure}
    \begin{subfigure}{.45\columnwidth}
    \includegraphics[width=1\columnwidth]{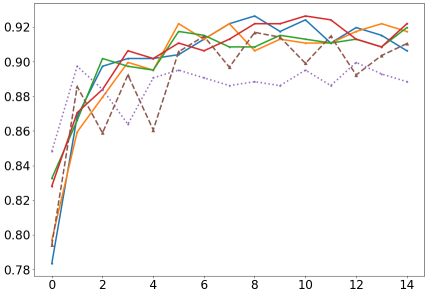}
    \subcaption{Accu. of BiLSTM}
    \end{subfigure}
    \begin{subfigure}{.45\columnwidth}
    \includegraphics[width=1\columnwidth]{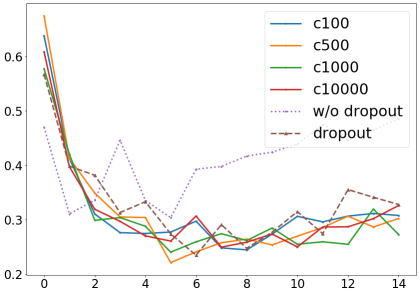}
    \subcaption{Loss of BiLSTM}
    \end{subfigure}    
    \caption{Visualization of training process deploying Dropout and Leaky Dropout with $c \in [10, 500, 1000, 10000]$ on TREC dataset. The $x$-axis of figures represents epochs of training.}
    \label{fig_accu_dropout}
\end{figure}

\subsubsection{Effect of Leaky Dropout}
To compare the vanilla Dropout with Leaky Dropout, we apply both mechanisms with AGA-CNN, AGA-LSTM, TextCNN and BiLSTM on TREC dataset and visualize the curves of Accuracy and Loss during the training in Figure \ref{fig_accu_dropout}. As shown in the figure, the Leaky Dropout generally produces a better performance when $c=500$. Specifically, the models with Leaky Dropout achieve a higher accuracy and show better convergence than the models with vanilla Dropout and without Dropout on the test set, which means the generalization ability of the model improves. Furthermore, the tail sections of curves reveal that Leaky Dropout is more robust to overfitting issue compared with conventional Dropout as there is severe fluctuation on curves of both accuracy and loss for models with conventional Dropout. Also, as shown in Figure \ref{fig_a} and \ref{fig_c}, AGA-based models with the Leaky Dropout achieve a higher accuracy that those with the conventional Dropout, which naturally makes sense since leaky mechanism can preserve infused information while the conventional Dropout may drop the neurons enriched by GI information.


\section{Related Work}
\subsection{Text classification}
Existing approaches employ deep architecture for supervised text classification and can achieve remarkable performance. Kim \cite{kim2014convolutional} proposes a classic TextCNN model for text classification, which significantly improved the accuracy of the classification task compared with machine learning approach. Johnson and Zhang \cite{johnson-zhang-2017-deep} build a deep architecture in pyramid shape with shortcut technique to extract dependency within a longer sequence. Zhang \emph{et al.} \cite{NIPS2015_5782} apply CNN to model character-level features and achieve competitive performance. Socher \emph{et al.} \cite{socher-etal-2013-recursive} use recursive neural networks explicitly exploiting time-series features. After that, several variants of the recurrent model are proposed, including Bi-LSTM \cite{graves2013hybrid} and GRU \cite{bahdanau2014neural} with more complex gate mechanisms. Zhou \emph{et al.} \cite{zhou2015c} present C-LSTM by joining both convolutional model and recurrent model to utilize sequential dependency upon local temporal features. Yang \emph{et al.} \cite{yang-etal-2016-hierarchical} propose the Hierarchical Attention Network to imitate the hierarchical structure of sentences and capture both word- and sentence-level features. These works mainly focus on architecture design for better feature extraction and selection, while our work coalesces semantic features with additional information and highlights the design of the fusion mechanism.

\subsection{Classifier with additional knowledge}
There is a large body of relevant literature to enhance classification performance using external knowledge in NLP.
Researchers create and exploit many active features incorporating information from various domains, including but not limited to linguistics, psychology and knowledge base. Post and Bergsma \cite{post2013explicit} utilize syntactic structure features such as POS tagging and dependency parsing to improve the performance of classification. Teng \emph{et al.} \cite{teng2016context} and Liang \emph{et al.} \cite{liang2018univ} fuse emotional lexicon into the model framework for sentiment analysis. Chen \emph{et al.} \cite{Chen2019DeepST} introduce conceptual information and entity links from knowledge base into the model pipeline through attention mechanism. Wang \emph{et al.} \cite{ijcai2017406} conceptualize sentence as a set of concepts using taxonomy knowledge base and obtain the embeddings by merging concepts on top of pre-trained word vectors, which can capture ampler contextual information facilitated by deep models. These works fail to concern the necessity and compatibility of added information, which probably bring more noise to the original semantic features and increase the cost of computation. \par

\section{Conclusion \& Future work}
In this paper, we propose the Adaptive Gate Attention module to incorporate global statistical features and conduct extensive experiments with both CNN-based framework and RNN-based framework to show the effectiveness of the proposed method. The proposed AGA module can merge necessary global information only while preserving essential semantic features, in which we provide a deep insight into the framework design of introducing additional knowledge. Moreover, the AGA module has great flexibility in use and can be extended to various relevant works. We also propose a novel Leaky Dropout mechanism to enhance the model generalization ability to enhance the model generalization ability and conduct additional experiments to demonstrate its effectiveness. Due to the page limit, we cannot do a comprehensive review with complete experiments on the Leaky Dropout, so we plan to examine the Leaky Dropout with theoretical analysis in our future work.

\bibliographystyle{named}
\bibliography{AGA_GI}

\begin{thebibliography}{}

\bibitem[\protect\citeauthoryear{Bahdanau \bgroup \em et al.\egroup
  }{2014}]{bahdanau2014neural}
Dzmitry Bahdanau, Kyunghyun Cho, and Yoshua Bengio.
\newblock Neural machine translation by jointly learning to align and
  translate.
\newblock {\em arXiv preprint arXiv:1409.0473}, 2014.

\bibitem[\protect\citeauthoryear{Chen \bgroup \em et al.\egroup
  }{2019}]{Chen2019DeepST}
Jindong Chen, Yizhou Hu, Jingping Liu, Yanghua Xiao, and Haiyun Jiang.
\newblock Deep short text classification with knowledge powered attention.
\newblock In {\em AAAI}, 2019.

\bibitem[\protect\citeauthoryear{Graves \bgroup \em et al.\egroup
  }{2013}]{graves2013hybrid}
Alex Graves, Navdeep Jaitly, and Abdel-rahman Mohamed.
\newblock Hybrid speech recognition with deep bidirectional lstm.
\newblock In {\em 2013 IEEE Workshop on Automatic Speech Recognition and
  Understanding}, pages 273--278, Dec 2013.

\bibitem[\protect\citeauthoryear{Hochreiter and
  Schmidhuber}{1997}]{hochreiter1997}
Sepp Hochreiter and J\"{u}rgen Schmidhuber.
\newblock Long short-term memory.
\newblock {\em Neural Comput.}, 9(8):1735–1780, November 1997.

\bibitem[\protect\citeauthoryear{Hu and Liu}{2004}]{hu2004mining}
Minqing Hu and Bing Liu.
\newblock Mining and summarizing customer reviews.
\newblock In {\em Proceedings of the tenth ACM SIGKDD international conference
  on Knowledge discovery and data mining}, pages 168--177. ACM, 2004.

\bibitem[\protect\citeauthoryear{Johnson and
  Zhang}{2017}]{johnson-zhang-2017-deep}
Rie Johnson and Tong Zhang.
\newblock Deep pyramid convolutional neural networks for text categorization.
\newblock In {\em Proceedings of the 55th Annual Meeting of the Association for
  Computational Linguistics (Volume 1: Long Papers)}, pages 562--570.
  Association for Computational Linguistics, July 2017.

\bibitem[\protect\citeauthoryear{Joulin \bgroup \em et al.\egroup
  }{2017}]{joulin-etal-2017-bag}
Armand Joulin, Edouard Grave, Piotr Bojanowski, and Tomas Mikolov.
\newblock Bag of tricks for efficient text classification.
\newblock In {\em Proceedings of the 15th Conference of the {E}uropean Chapter
  of the Association for Computational Linguistics: Volume 2, Short Papers},
  pages 427--431. Association for Computational Linguistics, April 2017.

\bibitem[\protect\citeauthoryear{Kim}{2014}]{kim2014convolutional}
Yoon Kim.
\newblock Convolutional neural networks for sentence classification.
\newblock In {\em Proceedings of the 2014 Conference on Empirical Methods in
  Natural Language Processing ({EMNLP})}, pages 1746--1751. Association for
  Computational Linguistics, October 2014.

\bibitem[\protect\citeauthoryear{Li and Roth}{2002}]{li2002learning}
Xin Li and Dan Roth.
\newblock Learning question classifiers.
\newblock In {\em Proceedings of the 19th international conference on
  Computational linguistics-Volume 1}, pages 1--7. Association for
  Computational Linguistics, 2002.

\bibitem[\protect\citeauthoryear{Liang \bgroup \em et al.\egroup
  }{2018}]{liang2018univ}
Weiming Liang, Haoran Xie, Yanghui Rao, Raymond~Y.K. Lau, and Fu~Lee Wang.
\newblock Universal affective model for readers’ emotion classification over
  short texts.
\newblock {\em Expert Systems with Applications}, 114:322 -- 333, 2018.

\bibitem[\protect\citeauthoryear{Mikolov \bgroup \em et al.\egroup
  }{2018}]{mikolov2018advances}
Tomas Mikolov, Edouard Grave, Piotr Bojanowski, Christian Puhrsch, and Armand
  Joulin.
\newblock Advances in pre-training distributed word representations.
\newblock In {\em Proceedings of the International Conference on Language
  Resources and Evaluation (LREC 2018)}, 2018.

\bibitem[\protect\citeauthoryear{Pang and Lee}{2004}]{pang2004sentimental}
Bo~Pang and Lillian Lee.
\newblock A sentimental education: Sentiment analysis using subjectivity
  summarization based on minimum cuts.
\newblock In {\em Proceedings of the 42nd annual meeting on Association for
  Computational Linguistics}, page 271. Association for Computational
  Linguistics, 2004.

\bibitem[\protect\citeauthoryear{Post and Bergsma}{2013}]{post2013explicit}
Matt Post and Shane Bergsma.
\newblock Explicit and implicit syntactic features for text classification.
\newblock In {\em Proceedings of the 51st Annual Meeting of the Association for
  Computational Linguistics (Volume 2: Short Papers)}, pages 866--872, 2013.

\bibitem[\protect\citeauthoryear{Socher \bgroup \em et al.\egroup
  }{2013a}]{socher2013recursive}
Richard Socher, Alex Perelygin, Jean Wu, Jason Chuang, Christopher~D Manning,
  Andrew Ng, and Christopher Potts.
\newblock Recursive deep models for semantic compositionality over a sentiment
  treebank.
\newblock In {\em Proceedings of the 2013 conference on empirical methods in
  natural language processing}, pages 1631--1642, 2013.

\bibitem[\protect\citeauthoryear{Socher \bgroup \em et al.\egroup
  }{2013b}]{socher-etal-2013-recursive}
Richard Socher, Alex Perelygin, Jean Wu, Jason Chuang, Christopher~D. Manning,
  Andrew Ng, and Christopher Potts.
\newblock Recursive deep models for semantic compositionality over a sentiment
  treebank.
\newblock In {\em Proceedings of the 2013 Conference on Empirical Methods in
  Natural Language Processing}, pages 1631--1642. Association for Computational
  Linguistics, October 2013.

\bibitem[\protect\citeauthoryear{Srivastava \bgroup \em et al.\egroup
  }{2014}]{srivastava2014dropout}
Nitish Srivastava, Geoffrey Hinton, Alex Krizhevsky, Ilya Sutskever, and Ruslan
  Salakhutdinov.
\newblock Dropout: a simple way to prevent neural networks from overfitting.
\newblock {\em The journal of machine learning research}, 15(1):1929--1958,
  2014.

\bibitem[\protect\citeauthoryear{Teng \bgroup \em et al.\egroup
  }{2016}]{teng2016context}
Zhiyang Teng, Duy-Tin Vo, and Yue Zhang.
\newblock Context-sensitive lexicon features for neural sentiment analysis.
\newblock In {\em Proceedings of the 2016 Conference on Empirical Methods in
  Natural Language Processing}, pages 1629--1638. Association for Computational
  Linguistics, November 2016.

\bibitem[\protect\citeauthoryear{Wang \bgroup \em et al.\egroup
  }{2017}]{ijcai2017406}
Jin Wang, Zhongyuan Wang, Dawei Zhang, and Jun Yan.
\newblock Combining knowledge with deep convolutional neural networks for short
  text classification.
\newblock In {\em Proceedings of the Twenty-Sixth International Joint
  Conference on Artificial Intelligence, {IJCAI-17}}, pages 2915--2921, 2017.

\bibitem[\protect\citeauthoryear{Wiebe \bgroup \em et al.\egroup
  }{2005}]{wiebe2005annotating}
Janyce Wiebe, Theresa Wilson, and Claire Cardie.
\newblock Annotating expressions of opinions and emotions in language.
\newblock {\em Language resources and evaluation}, 39(2-3):165--210, 2005.

\bibitem[\protect\citeauthoryear{Yang \bgroup \em et al.\egroup
  }{2016}]{yang-etal-2016-hierarchical}
Zichao Yang, Diyi Yang, Chris Dyer, Xiaodong He, Alex Smola, and Eduard Hovy.
\newblock Hierarchical attention networks for document classification.
\newblock In {\em Proceedings of the 2016 Conference of the North {A}merican
  Chapter of the Association for Computational Linguistics: Human Language
  Technologies}, pages 1480--1489. Association for Computational Linguistics,
  June 2016.

\bibitem[\protect\citeauthoryear{Zhang \bgroup \em et al.\egroup
  }{2015}]{NIPS2015_5782}
Xiang Zhang, Junbo Zhao, and Yann LeCun.
\newblock Character-level convolutional networks for text classification.
\newblock In {\em Advances in Neural Information Processing Systems 28}, pages
  649--657. Curran Associates, Inc., 2015.

\bibitem[\protect\citeauthoryear{Zhou \bgroup \em et al.\egroup
  }{2015}]{zhou2015c}
Chunting Zhou, Chonglin Sun, Zhiyuan Liu, and Francis Lau.
\newblock A c-lstm neural network for text classification.
\newblock {\em arXiv preprint arXiv:1511.08630}, 2015.

\end{thebibliography}

\end{document}